\documentclass[dvipsnames]{article}

\usepackage{colm2024_conference}

\usepackage{booktabs}
\usepackage{graphicx}
\usepackage{enumitem}
\usepackage{float}
\usepackage{placeins}
\usepackage{microtype}
\usepackage{amsmath}
\usepackage{fontspec}
\usepackage{xeCJK}
\setCJKsansfont[BoldFont=FandolHei-Bold.otf]{FandolHei-Regular.otf}
\setCJKmonofont{FandolFang-Regular.otf}
\usepackage[russian,english]{babel}
\definecolor{lightgray}{rgb}{0.9,0.9,0.9}
\usepackage{caption}
\usepackage{subcaption}
\usepackage{url}
\usepackage{multirow}
\usepackage{multicol}
\usepackage{tabularx}
\usepackage{makecell}
\usepackage[raster,skins]{tcolorbox}
\usepackage{adjustbox}
\usepackage{xurl}
\usepackage[normalem]{ulem}
\useunder{\uline}{\ul}{}
\usepackage{pifont}
\usepackage{xcolor}
\usepackage{array}
\newcommand{\checkmark}{\ding{51}}

\definecolor{qianfan_blue}{HTML}{0068B7}
\definecolor{qianfan_lightblue}{HTML}{00A0E9}
\definecolor{qianfan_gray}{HTML}{7F7F7F}

\title{Qianfan-OCR: A Unified End-to-End Model for Document Intelligence}

\author{
Baidu Qianfan Team
}

\begin{document}

\maketitle

\begin{abstract}
\emergencystretch=1em
  We present Qianfan-OCR, a 4B-parameter end-to-end document intelligence model that unifies document parsing, layout analysis, and document understanding within a single vision-language architecture. Unlike traditional multi-stage OCR pipelines that chain separate layout detection, text recognition, and language comprehension modules, Qianfan-OCR performs direct image-to-Markdown conversion and supports a broad range of prompt-driven tasks -- from structured document parsing and table extraction to chart understanding, document question answering, and key information extraction -- all within one model.
  A practical limitation of end-to-end OCR is the loss of explicit layout analysis, a capability that pipeline users routinely rely on for element localization and type classification. We introduce Layout-as-Thought to bridge this gap: an optional thinking phase triggered by $\langle$think$\rangle$ tokens, where the model generates structured layout representations (bounding boxes, element types, and reading order) before producing final outputs. This mechanism serves two purposes: (1) it recovers layout analysis functionality within the end-to-end paradigm, enabling users to obtain spatial grounding results directly; and (2) it provides targeted accuracy improvements on documents with complex layouts, cluttered elements, or non-standard reading orders, where structural priors help resolve recognition ambiguities.
  On OCR-specific benchmarks, Qianfan-OCR ranks first among all end-to-end models on OmniDocBench v1.5 (93.12) and OlmOCR Bench (79.8). It also achieves strong results on general OCR benchmarks including OCRBench (880), OCRBenchv2, and CCOCR, as well as document understanding tasks such as DocVQA, ChartQA, and CharXiv, matching general vision-language models of comparable scale. On public Key Information Extraction benchmarks, Qianfan-OCR achieves the highest average score, surpassing Gemini-3.1-Pro, Gemini-3-Pro, Seed-2.0, and Qwen3-VL-235B-A22B. The model is publicly accessible through Baidu AI Cloud Qianfan platform, with usage examples and best practices available at \url{https://github.com/baidubce/Qianfan-VL}.
\end{abstract}

\section{Introduction}

Current OCR systems face a three-way trade-off between cost, accuracy, and capability.
\textit{Traditional OCR pipelines} based on small specialized models offer low inference cost and high throughput, but require complex multi-stage preprocessing and postprocessing to handle diverse document layouts.
\textit{Specialized OCR large models}~\citep{wei2024got, wei2025deepseek, cui2025paddleocrvl, poznanski2025olmocr} improve accuracy through two-stage architectures -- layout detection followed by element-wise recognition -- but introduce deployment complexity, inter-stage error propagation, and irreversible loss of visual context during text extraction.
\textit{General vision-language models}~\citep{liu2024visual, chen2024internvl} offer broad multimodal capabilities but incur higher inference costs and underperform specialized systems on structured document parsing tasks requiring precise layout preservation.

In practice, industrial OCR applications -- document retrieval with chunking and indexing, contract review, key information extraction from receipts and certificates -- often chain detection models, OCR models, and separate LLMs for downstream understanding. This fragmented approach increases deployment cost, limits end-to-end optimization, and requires careful orchestration of heterogeneous components.

We introduce Qianfan-OCR, a 4B-parameter unified end-to-end model that addresses these limitations with three key designs:

\textbf{End-to-End Architecture:}
Qianfan-OCR integrates layout analysis, text recognition, and semantic understanding into a single vision-language model, eliminating inter-stage error propagation and enabling joint optimization across all tasks. The end-to-end design allows the model to retain full visual context throughout processing -- spatial relationships, chart structures, and formatting that pipeline systems discard during text extraction. For tasks that do not require explicit layout analysis (e.g., simple document transcription or scene text recognition), the model directly outputs results without mandatory layout preprocessing.

\textbf{Layout-as-Thought:}
A practical limitation of end-to-end OCR is the loss of explicit layout analysis -- a capability that pipeline systems inherently provide through dedicated detection modules. Layout-as-Thought recovers this within the end-to-end paradigm: an optional thinking phase triggered by $\langle$think$\rangle$ tokens, where the model generates bounding boxes, element types, and reading order before producing final outputs. This serves two purposes: (1) \textit{functional} -- users obtain structured layout results (element localization, type classification, spatial grounding) directly from an end-to-end model, bridging a key functionality gap relative to pipeline systems; (2) \textit{enhancement} -- the explicit structural priors help resolve ambiguities in documents with cluttered elements, complex multi-column layouts, or non-standard reading orders. For well-structured documents where the model already performs well, the layout phase is unnecessary; it targets the subset of challenging cases where structural reasoning provides measurable gains.

\textbf{Unified OCR and Understanding:}
Beyond conventional OCR tasks (document parsing, handwriting recognition, table extraction), Qianfan-OCR handles cognitively demanding tasks including chart understanding, document question answering, and key information extraction -- tasks requiring both precise text perception and semantic reasoning. Traditional OCR models lack comprehension capabilities, limiting them to character-level extraction; general VLMs possess reasoning abilities but underperform on structured parsing. Qianfan-OCR bridges this divide, combining OCR-specialist-level accuracy with document understanding capabilities in a single model controllable through prompts.

\begin{figure*}[t]
    \centering
    \includegraphics[width=0.98\textwidth]{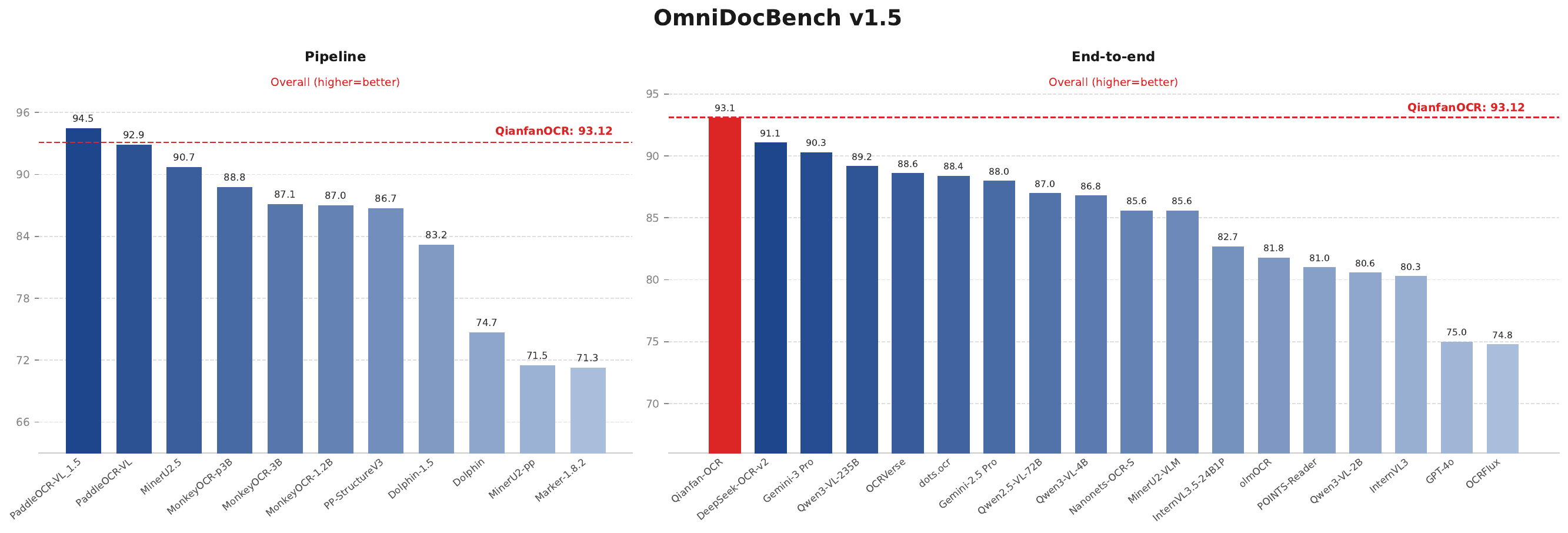}
    \caption{Performance on OmniDocBench v1.5 across pipeline (left) and end-to-end (right) models. Qianfan-OCR (red) achieves 93.12, ranking first among all end-to-end models. The red dashed line indicates Qianfan-OCR's score for cross-category comparison.}
    \label{fig:benchmarks}
\end{figure*}

\begin{figure*}[t]
    \centering
    \includegraphics[width=0.98\textwidth]{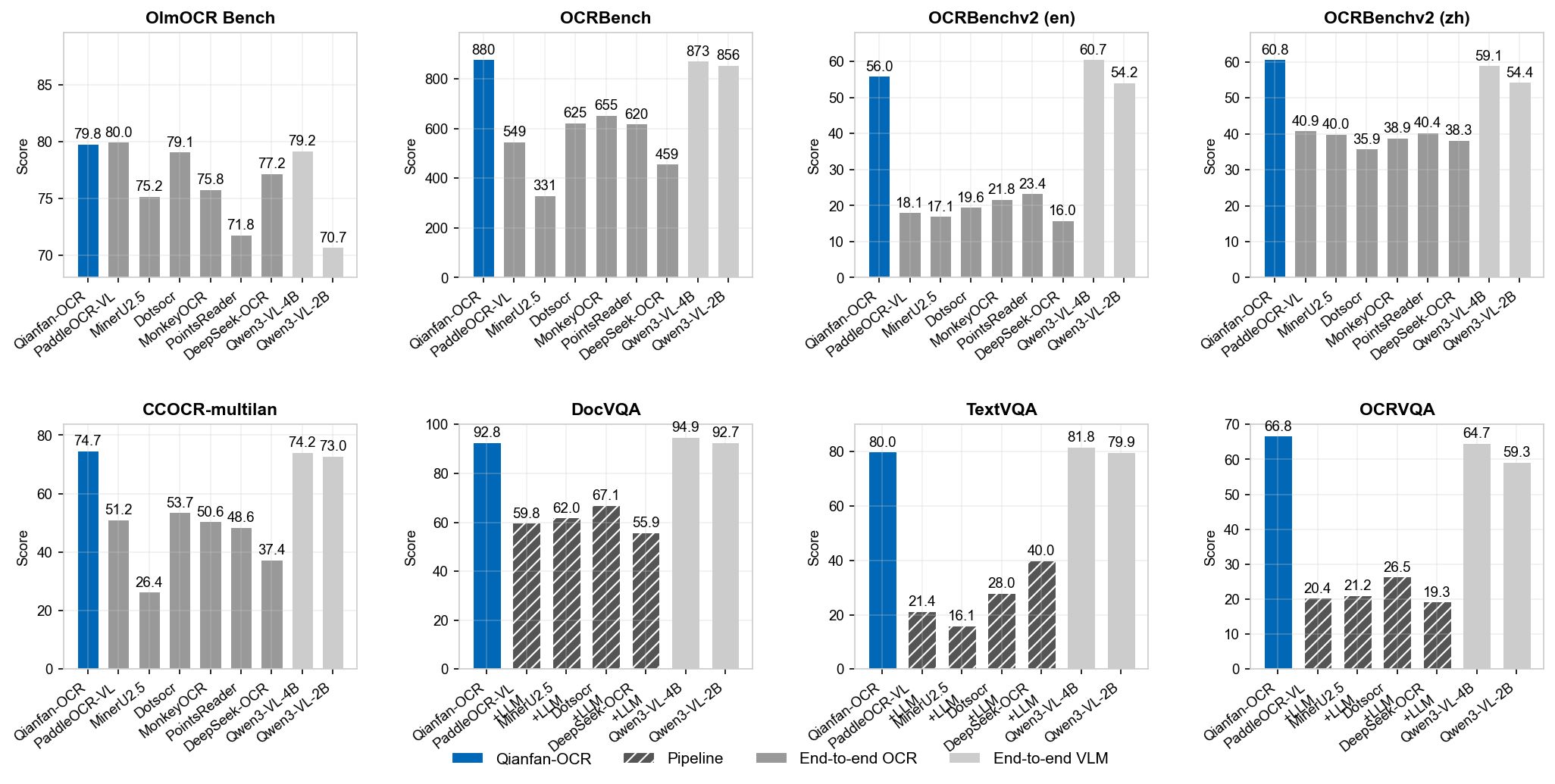}
    \caption{Performance on general OCR and document understanding benchmarks. \textbf{Top row:} OlmOCR Bench, OCRBench, OCRBenchv2 (en), and OCRBenchv2 (zh). \textbf{Bottom row:} CCOCR-multilan, and document understanding tasks (DocVQA, TextVQA, OCRVQA) where two-stage OCR+LLM pipelines (hatched bars) show significant degradation compared to end-to-end models.}
    \label{fig:other_benchmarks}
\end{figure*}

\section{Related Work}

We review three technical routes in OCR and position Qianfan-OCR relative to each.

\textbf{Pipeline OCR Systems.}
Pipeline systems~\cite{cui2025paddleocr} decompose document parsing into layout detection, element-wise recognition, and rule-based assembly. Recent systems such as PaddleOCR-VL~\cite{cui2025paddleocrvl}, MonkeyOCR, and MinerU 2.5 pair lightweight detection models with VLM-based recognizers, achieving strong accuracy with modular efficiency. Their key advantage is explicit layout analysis output (bounding boxes, element types), but they suffer from inter-stage error propagation and irreversible loss of visual context during text extraction. Qianfan-OCR recovers the layout analysis capability through Layout-as-Thought while avoiding the pipeline's cascading error problem.

\textbf{End-to-End OCR Models.}
End-to-end approaches directly map document images to structured outputs. Nougat~\cite{blecher2023nougat} demonstrated feasibility on academic papers; GOT-OCR 2.0~\cite{wei2024got} broadened format support (Markdown, LaTeX, TikZ) at 580M parameters; DeepSeek-OCR~\cite{wei2025deepseek} introduced context optical compression for efficiency; olmOCR~\cite{poznanski2025olmocr} scaled SFT-based training on large-scale web documents, while its successor olmOCR 2 further introduced GRPO reinforcement learning with unit-test rewards. More recently, Dolphin v2 proposed analyze-then-parse, and Logics-Parsing and Infinity-Parser explored layout-aware RL for structure prediction. These models primarily focus on recognition accuracy or efficiency but lack explicit layout analysis output -- a functionality gap that Qianfan-OCR's Layout-as-Thought addresses. Qianfan-OCR relies on supervised fine-tuning with high-quality layout annotations, a complementary paradigm that future work could augment with reinforcement learning.

\textbf{General Vision-Language Models.}
Large VLMs such as Qwen-VL~\cite{bai2023qwen,bai2025qwen2}, InternVL~\cite{chen2024internvl,zhu2025internvl3}, and Gemini exhibit OCR capabilities as a byproduct of broad multimodal training, but are not optimized for structured document parsing: they incur higher inference costs, lack fine-grained layout control, and underperform specialized systems on structure-sensitive metrics (e.g., table TEDS, reading order accuracy). Qianfan-OCR targets OCR-specialist-level accuracy at comparable inference cost to these models, while additionally supporting explicit layout analysis and prompt-driven task control.

\begin{figure}[t]
    \centering
    \includegraphics[width=0.95\textwidth]{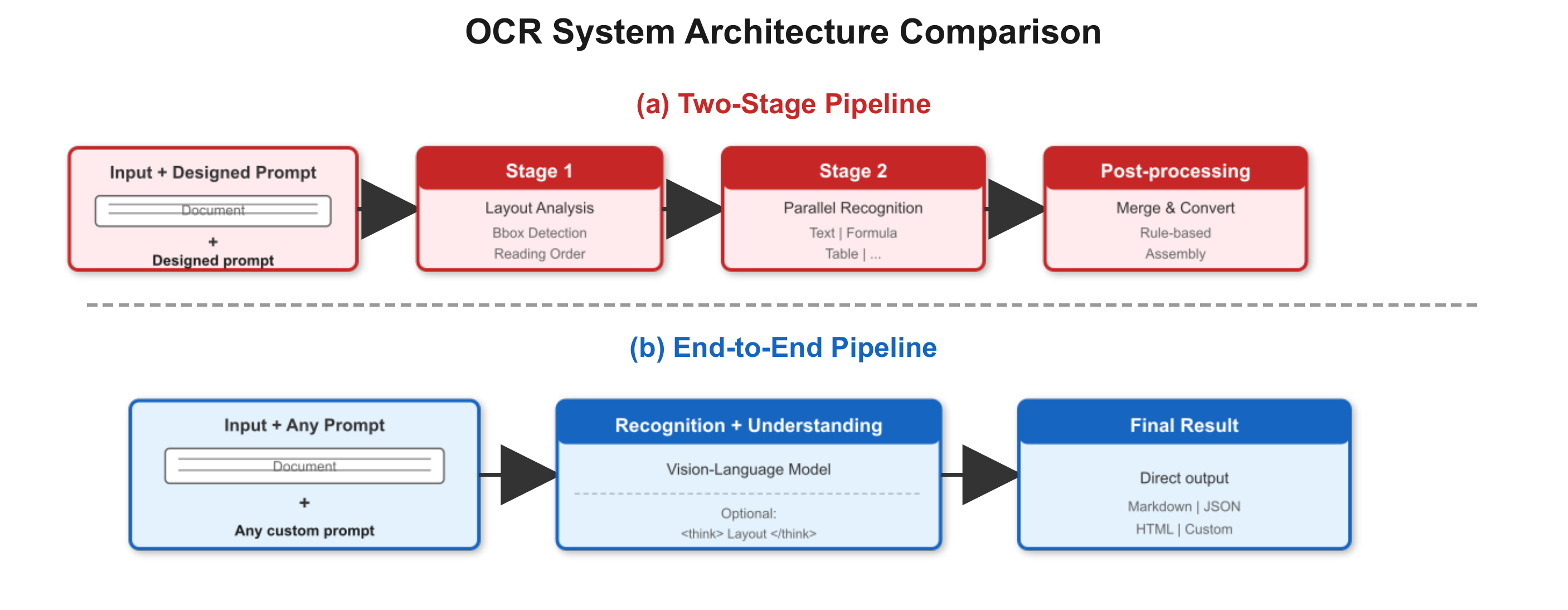}
    \caption{Architectural comparison between traditional two-stage OCR pipeline and Qianfan-OCR's end-to-end approach. \textbf{(a)} Traditional pipeline systems separate layout analysis and content recognition into independent stages, suffering from error propagation and irreversible loss of visual context. \textbf{(b)} Qianfan-OCR unifies all processing into a single vision-language model, accepting custom prompts for flexible task control and optionally generating intermediate layout reasoning via Layout-as-Thought ($\langle$think$\rangle$ tokens).}
    \label{fig:ocr_comparison}
\end{figure}

\section{Model Architecture and Training Data}
\label{sec:architecture}

\subsection{Architecture Overview}

Qianfan-OCR adopts the multimodal bridging architecture from Qianfan-VL~\citep{dong2025qianfanvl}, consisting of three core components: a vision encoder for flexible visual encoding, a lightweight projection adapter for cross-modal alignment, and a language model backbone for text generation and reasoning. The overall architecture is illustrated in Figure~\ref{fig:ocr_comparison}(b).

\textbf{Vision Encoder.}
The vision encoder employs \textbf{Qianfan-ViT}, pretrained as part of the Qianfan-VL framework~\citep{dong2025qianfanvl}. It adopts the AnyResolution design that dynamically tiles input images into 448$\times$448 patches, supporting variable-resolution inputs up to 4K. This is critical for OCR tasks where documents contain dense text, small fonts, and complex layouts that require high-resolution processing. The encoder consists of 24 Transformer layers with 1024 hidden dimensions, 16 attention heads, and a 14$\times$14 patch size, producing 256 visual tokens per tile. With a maximum of 16 tiles, the encoder can represent a single document image with up to 4,096 visual tokens, providing sufficient spatial resolution for fine-grained character recognition.

\textbf{Language Model Backbone.}
We adopt \textbf{Qwen3-4B}~\citep{bai2025qwen2} as the language model backbone. The model has 4.0B total parameters (3.6B non-embedding), 36 layers, 2560 hidden dimensions, and a 32K native context window (extendable to 131K via YaRN). This scale strikes a balance between reasoning capability and deployment efficiency -- large enough for complex document understanding and layout reasoning, yet practical for single-GPU serving in production. The model uses Grouped-Query Attention (GQA)~\citep{ainslie2023gqa} with 32 query heads and 8 KV heads, reducing KV cache memory by 4$\times$ compared to standard multi-head attention while maintaining generation quality. RMSNorm~\citep{zhang2019root} is used for layer normalization, improving training stability.

\textbf{Cross-Modal Adapter.}
A lightweight two-layer MLP with GELU activation bridges the vision encoder and the language model, projecting visual features from the encoder's representation space (1024 dimensions) into the language model's embedding space (2560 dimensions). This simple design minimizes adapter parameters while ensuring effective cross-modal alignment. During Stage~1 training, only the adapter is trained with a higher learning rate for fast alignment, while subsequent stages perform full-parameter training.

\subsection{Large-Scale Data Synthesis Pipelines}

We develop six data synthesis pipelines covering document parsing, key information extraction, complex tables, chart understanding, formula recognition, and multilingual OCR.

\textbf{Document Parsing Data Synthesis:}
We construct an automated pipeline that converts document images into structured Markdown using PaddleOCR-VL~\citep{cui2025paddleocrvl} for layout detection and content recognition, with bbox coordinates normalized to [0, 999] for resolution invariance. Tables are converted to HTML via OTSL intermediate format, and formulas are wrapped in \$\$ blocks. Automatic filtering removes repetitive or extreme-length samples, and image-level augmentations (compression, flipping, blur) improve robustness.

A key design choice is the layout label system. We compare PaddleOCR-VL and MinerU~2.5 along label granularity and detection accuracy. The main difference lies in body text labels: PaddleOCR-VL provides fine-grained categories (\texttt{text}, \texttt{vertical\_text}, \texttt{paragraph\_title}, \texttt{doc\_title}, \texttt{abstract}, \texttt{content}, \texttt{reference}, \texttt{reference\_content}, \texttt{aside\_text}), while MinerU~2.5 uses coarser labels (\texttt{text}, \texttt{title}, \texttt{list}, \texttt{aside\_text}). Finer granularity directly benefits downstream tasks -- e.g., distinguishing \texttt{abstract} from \texttt{content} supports structured extraction from papers, and separating \texttt{reference} enables clean bibliography parsing. We also evaluate both systems on a multi-type document layout benchmark, where PaddleOCR-VL achieves consistently higher detection accuracy. We therefore adopt PaddleOCR-VL's label system and use it as the annotation engine. Our final taxonomy contains 25 categories in four groups: \textit{text elements} (12 labels), \textit{headers/footers} (4), \textit{figures/tables} (6), and \textit{formulas} (3).

\textbf{Layout-as-Thought Data Construction:}
We construct training data where the model generates structured layout analysis within \texttt{<think>} tokens before final output, listing bbox coordinates, element labels, and content summaries as intermediate reasoning enclosed in \texttt{<layout>...</layout>} tags. Users activate this by appending \texttt{<think>} tokens to queries. The layout phase focuses model attention on relevant document regions before generation, improving performance on documents requiring spatial reasoning (complex layouts, multi-column text, interleaved figures).

\textbf{Key Information Extraction (KIE):}
For KIE tasks, we construct datasets for two scenarios: complete extraction ("what you see is what you get") and targeted extraction (user-specified keys). To address hallucination in teacher models, we combine open-source data with small model pre-annotations for multi-model collaborative labeling. We implement semantic generalization for keys across different regions and formats, constructing multiple synonymous descriptions for the same field. The pipeline includes quality enhancement through direction correction and image enhancement for low-resolution inputs, hard rule filtering using business logic (e.g., verifying "unit price × quantity = total"), and difficult sample mining for long sequences with 5+ detail rows and dense text documents. Sample distribution is rebalanced based on task difficulty to enhance stability in extreme scenarios.

\textbf{Complex Tables:}
We combine programmatic synthesis with real document extraction. The programmatic pipeline randomly generates tables with 3-20 rows/columns supporting random cell merging, populates content via Faker library or LLMs covering diverse data types, randomly samples from 50+ professional CSS themes, renders via Jinja2 and KaTeX engines, and applies geometric transformations, color perturbations, and blur augmentations. For real document tables, we use internal parsing tools to detect and extract table regions, parse with both PaddleOCR-VL and internal table models, convert both outputs to HTML, and perform consistency validation to filter samples with significant structural or content differences, ensuring reliable annotations while preserving real document layout and noise characteristics.

\textbf{Chart Understanding:}
We build an automated synthesis pipeline based on arXiv LaTeX sources (2022-present). The pipeline systematically extracts Figure code blocks through rule parsing, re-renders using TexLive engine to obtain lossless vector images, leverages caption parameters as ground truth, and uses VLMs to generate detailed visual descriptions capturing visual encoding, statistical features, spatial layout, and fine-grained distribution characteristics. We categorize 11 mainstream chart types and construct "metadata + visual description" driven synthesis. Custom reasoning tasks are designed for different chart types: trend analysis for line charts, correlation and distribution features for scatter plots, outlier detection for box plots. This pipeline synthesizes over 300,000 high-accuracy, diverse multimodal instruction-tuning samples.

\textbf{Multilingual OCR Data Construction:}
To extend language coverage to 192 languages, we adopt a \textit{reverse synthesis} approach starting from the HPLT multilingual corpus. The pipeline performs text-font renderability filtering using fonttools character set validation, then renders document images with differentiated handling for different writing systems (Latin, Cyrillic, Arabic, South Asian, Southeast Asian, Han). Key features include automatic RTL text direction detection, Arabic character reshaping, and word-level line breaking. Diverse typesetting variations (font size, column layout, margins, spacing, texture backgrounds) are randomized to approximate real document distributions.

\textbf{Document Image Augmentation:}
We employ two augmentation pipelines: one for OCR tasks (allowing mild geometric perturbations) and one for layout parsing tasks (preserving geometric consistency). Both apply three noise stages: (1) \textit{text noise} (broken strokes, ink bleeding, character misalignment), (2) \textit{background noise} (texture, color drift, watermarks), and (3) \textit{imaging noise} (blur, moiré, shadows, exposure variation). Additionally, rotation augmentation (90\textdegree, 180\textdegree, 270\textdegree, and $\pm$15\textdegree) significantly improves performance on KIE and table recognition tasks where documents frequently appear in non-standard orientations.

Through these specialized synthesis pipelines, we generate large-scale, high-quality training data covering diverse OCR scenarios, providing comprehensive data support for Qianfan-OCR's multi-stage progressive training.

\section{Training Recipe}
\label{sec:training}

Qianfan-OCR adopts the proven multi-stage progressive training methodology from Qianfan-VL~\citep{dong2025qianfanvl}, which systematically builds model capabilities from basic cross-modal alignment through advanced reasoning tasks.
The key adaptation for OCR scenarios lies in the \textbf{data mixture composition}, where we significantly enhance OCR-specific domains while maintaining the overall training framework.
The training pipeline consists of four stages with carefully designed data distributions optimized for document intelligence.

\textbf{Stage 1: Cross-Modal Alignment (50B tokens)} -- Establishes fundamental vision-language alignment with adapter-only training, using basic image-caption pairs and simple OCR tasks to ensure stable initialization.

\textbf{Stage 2: Foundational OCR Training (2T tokens)} -- Develops comprehensive OCR capabilities through full parameter training with OCR-heavy data mixture: Document OCR (45\%), Scene OCR (25\%), Caption (15\%), and Specialized OCR tasks including handwriting, formulas, tables, and multilingual text (15\%).

\textbf{Stage 3: Domain-Specific Enhancement (800B tokens)} -- Implements targeted enhancement for enterprise-critical OCR domains with balanced mixture: Complex Tables (22\%), Formula Recognition (20\%), Chart Understanding (18\%), Information Extraction (18\%), Multilingual OCR (12\%), and Document Understanding (10\%). Maintains 70\% domain-specific data and 30\% general data to enhance specialization while preventing catastrophic forgetting.

\textbf{Stage 4: Instruction Tuning and Reasoning Enhancement (millions of instruction samples)} -- Covers a comprehensive set of document intelligence tasks including document parsing, layout analysis, handwriting recognition, scene text recognition, formula recognition, table recognition, multi-page document parsing, chart QA, document QA, and complex table QA. The instruction data is constructed through three complementary strategies: (1)~\textit{Public data curation}: we collect publicly available OCR-related training datasets and perform instruction rewriting and generalization using DeepSeek models to diversify prompt styles and task formulations; (2)~\textit{Reverse synthesis}: for tasks amenable to reverse generation (e.g., tables, exam papers), we construct large-scale QA pairs by generating questions conditioned on structured ground-truth content; (3)~\textit{Chart data mining}: we extract chart-figure pairs from a large corpus of academic papers via their LaTeX source code and generate chart understanding QA pairs grounded in the original source, significantly enhancing chart comprehension capabilities. All instruction data undergoes systematic prompt generalization and rewriting to improve robustness to diverse user instructions.

The critical differentiator from general VLM training lies in the \textbf{OCR-centric data composition} throughout all stages, with particular emphasis on document parsing, table/chart understanding, and information extraction tasks.
Detailed data synthesis pipelines for each domain are described in Section~\ref{sec:architecture}.

\textbf{Training Infrastructure and Iteration Strategy.}
All training is conducted on \textbf{1,024 Baidu Kunlun P800 chips} using 3D parallelism (data, tensor, and pipeline parallelism with communication-computation overlap), processing over 2.85T tokens across all stages. In practice, Stages~1 and~2 are trained once to establish a stable foundation checkpoint, while Stages~3 and~4 are iterated multiple times to explore different domain-specific data mixtures, sampling ratios, and instruction tuning configurations. Since Stages~3 and~4 account for a smaller token budget (800B + instruction tuning vs.\ 2T for Stage~2), each iteration completes quickly. The full four-stage pipeline completes \textbf{within a week}, and Stage~3/4 iterations take approximately one day each, supporting systematic ablation and optimization of OCR-specific training recipes.

\subsection{Hyperparameters}

Table~\ref{tab:training_hyperparams} summarizes the key hyperparameters for each training stage.

\begin{table}[H]
\centering
\small
\setlength{\tabcolsep}{4pt}
\begin{tabular}{l|cccc}
\toprule
\textbf{Hyperparameter} & \textbf{Stage 1} & \textbf{Stage 2} & \textbf{Stage 3} & \textbf{Stage 4} \\
\midrule
Trainable modules & Adapter & All & All & All \\
Tokens & 50B & 2T & 800B & -- \\
Global batch size (samples) & 1024 & 2048 & 2048 & 512 \\
Peak learning rate & 1e-3 & 2e-5 & 1e-5 & 1e-5 \\
LR schedule & Cosine & Cosine & Cosine & Cosine \\
Warmup ratio & 0.01 & 0.01 & 0.01 & 0.03 \\
Optimizer & \multicolumn{4}{c}{AdamW ($\beta_1$=0.9, $\beta_2$=0.95)} \\
Weight decay & \multicolumn{4}{c}{0.05} \\
Max sequence length & \multicolumn{4}{c}{32768} \\
\bottomrule
\end{tabular}
\caption{Training hyperparameters for each stage. Stage~1 trains only the adapter with a higher learning rate for fast alignment. Stages~2--4 use full-parameter training with lower learning rates.}
\label{tab:training_hyperparams}
\end{table}

\subsection{Ablation Study: Multi-Stage Training Effectiveness}

Prior to training the 4B Qianfan-OCR model, we conduct low-cost ablation studies on Qianfan-VL-8B -- a model from the same architectural family that has undergone large-scale general-purpose continual pretraining -- to systematically validate our multi-stage training recipe. Starting from a Stage~1 (adapter-only alignment) checkpoint, we evaluate the contribution of each subsequent training stage to OCR performance. All configurations employ full-parameter training, which prior Qianfan-VL experiments have confirmed to be essential for effective domain transfer. Results are summarized in Table~\ref{tab:ablation}.

\begin{table}[H]
\centering
\small
\setlength{\tabcolsep}{5pt}
\begin{tabular}{cccc|c}
\toprule
\textbf{Stage 1} & \textbf{Stage 2} & \textbf{Stage 3} & \textbf{Stage 4} & \textbf{AVG} \\
\midrule
\checkmark & -- & -- & \checkmark & 71.37 \\
\checkmark & -- & OCR & \checkmark & 75.97 \\
\checkmark & -- & OCR + General & \checkmark & 80.07 \\
\checkmark & General & -- & \checkmark & 83.47 \\
\checkmark & General & OCR & \checkmark & 84.09 \\
\checkmark & General & OCR + General & \checkmark & \textbf{84.39} \\
\bottomrule
\end{tabular}
\caption{Ablation study on multi-stage training effectiveness using Qianfan-VL-8B. ``OCR'' denotes domain-specific OCR data, ``General'' denotes general-purpose data, and ``OCR + General'' denotes a 1:1 mixture. Stage~4 is instruction tuning with a fixed set of alignment samples across all configurations. Average accuracy is computed over multiple OCR benchmarks.}
\label{tab:ablation}
\end{table}

\textbf{Stage~2 (foundational pretraining) is essential.} Skipping Stage~2 and proceeding directly from Stage~1 to domain-specific Stage~3 training yields significantly lower performance. Even the best Stage~3-only configuration (OCR + General mixture, 80.07\%) falls substantially short of Stage~2 followed by Stage~4 alone (83.47\%), indicating that large-scale general-purpose pretraining provides a critical capability foundation that cannot be substituted by domain-specific data alone.

\textbf{Domain-specific enhancement benefits from general data mixing.} In Stage~3, a 1:1 mixture of OCR-specific and general-purpose data (80.07\%) consistently outperforms pure OCR data (75.97\%), suggesting that maintaining general capability during domain specialization acts as an effective regularizer and prevents overfitting to narrow OCR patterns. The same trend holds when Stage~2 is included: OCR + General in Stage~3 (84.39\%) outperforms OCR-only Stage~3 (84.09\%).

\textbf{The complete pipeline achieves optimal performance.} The full four-stage configuration (Stage~1 $\rightarrow$ Stage~2 General $\rightarrow$ Stage~3 OCR+General $\rightarrow$ Stage~4) achieves the highest accuracy of 84.39\%, representing a +13.02\% absolute improvement over the Stage~1+4 baseline. Notably, this configuration also surpasses Qwen2.5-VL-7B (79.30\%) by +5.09\%, despite the latter being a strong general-purpose VLM. In practice, we keep the Stage~4 data volume fixed and incorporate all additional domain-specific data augmentation into Stage~3, which proves to be the most effective strategy for scaling OCR capabilities without compromising the general abilities established in Stage~2.

These findings, obtained at relatively low cost on the 8B model, directly motivated the training recipe adopted for the 4B Qianfan-OCR. The consistency of improvements across training stages suggests that this progressive recipe generalizes across model scales within the same architectural family.

\section{Evaluation Framework}
\label{sec:evaluation_framework}

To comprehensively assess Qianfan-OCR's capabilities across the spectrum from specialized OCR to document understanding, we employ a multi-dimensional evaluation framework spanning four key categories:

\textbf{Specialized OCR Model Benchmarks:}
We evaluate OCR-specific capabilities using benchmarks designed for specialized document parsing systems: Omni-Doc-Bench v1.5~\citep{ouyang2024omnidocbench} for diverse PDF document parsing, OLMOCRBench~\citep{olmocr2024} for end-to-end document OCR evaluation, CCOCR for multilingual OCR recognition across diverse scripts, and BigDocs~\citep{rodriguez2024bigdocs} for large-scale document processing tasks.
These benchmarks assess the model's ability to achieve recognition accuracy competitive with specialized OCR systems.

\textbf{General OCR Capability Benchmarks:}
To evaluate general-purpose optical character recognition across diverse scenarios, we use OCRBench~\citep{chen2024ocrbench} and OCRBench v2, comprehensive benchmarks covering scene text, document text, handwriting recognition, formula recognition, and multilingual text across various real-world conditions.
These evaluations ensure the model maintains robust OCR performance beyond specialized document parsing scenarios.

\textbf{Document Understanding Benchmarks:}
We assess document comprehension and reasoning capabilities using a diverse set of understanding-oriented benchmarks: TextVQA~\citep{singh2019towards} for scene text question answering, DocVQA~\citep{mathew2021docvqa} for document visual question answering, CharXiv document question (CharXiv\_DQ) and reasoning question (CharXiv\_RQ) tasks~\citep{wang2024charxiv} for academic document understanding, ChartQA~\citep{masry2022chartqa} and ChartQAPro~\citep{masry2025chartqapro} for chart interpretation, and ChartBench for comprehensive chart reasoning evaluation.
These benchmarks evaluate the model's ability to perform high-level semantic understanding and reasoning over visual document content, including complex table analysis and chart interpretation.

\textbf{Key Information Extraction (KIE):}
We evaluate KIE capabilities across five public benchmarks: OCRBench KIE~\citep{chen2024ocrbench} (structured field extraction from receipts, invoices, and forms), OCRBenchv2 KIE in both English and Chinese (cross-lingual extraction across diverse document templates), CCOCR KIE (Chinese document field extraction including ID cards, business licenses, and financial documents), and Nanonets KIE (real-world invoice and receipt parsing, measured by F1 score). All scores are normalized to a 0--100 scale. Since specialized OCR models lack native KIE capabilities, we compare against commercial large models (Gemini-3.1-Pro, Gemini-3-Pro, Seed-2.0) and open-source VLMs (Qwen3-4B-VL, Qwen3-VL-235B-A22B).

\textbf{Comparative Evaluation Against Pipeline Systems:}
To fairly compare against traditional OCR-then-LLM approaches, we establish baseline systems that combine specialized OCR models with language models of comparable parameter count to Qianfan-OCR.
This comparative evaluation on understanding benchmarks isolates the architectural effects from parameter count differences, providing controlled evidence for the trade-offs between end-to-end and pipeline approaches.

\section{Experimental Results}
\label{sec:evaluation}

We conduct comprehensive evaluations across OCR-specific benchmarks, general OCR benchmarks, and document understanding tasks. Our evaluation framework is primarily based on VLMEvalKit~\citep{duan2024vlmevalkit} with modifications tailored to different benchmark requirements.

\textbf{Evaluation Methodology.} For OCR-specific benchmarks, we follow official metrics or adopt results from recent publications. Specifically, we reference metrics from the OmniDocBench official leaderboard~\citep{ouyang2024omnidocbench}, DeepSeek-OCR v2~\citep{wei2026deepseekocr2visualcausal}, and PaddleOCR-VL 1.5~\citep{cui2026paddleocrvl15} papers. For benchmarks with official evaluation scripts (e.g., OmniDocBench v1.5), we directly reuse their implementations.

\textbf{Model Categorization.} We categorize comparison models based on their architectural paradigms:
\begin{itemize}[noitemsep,topsep=0pt]
\item \textbf{Pipeline OCR Systems:} Models that perform layout analysis first, followed by parallel text recognition (e.g., PaddleOCR-VL, MonkeyOCR series, Dolphin series).
\item \textbf{End-to-end Models:} Vision-language models that directly process images through prompts without separate detection stages, including both specialized OCR models (e.g., MinerU2.5, Dotsocr, DeepSeek-OCR) and general VLMs (e.g., Qwen3-VL, GPT-4o, Gemini).
\end{itemize}

\textbf{Benchmark-Specific Considerations.} For general OCR benchmarks (OCRBench, OCRBenchv2, CCOCR-multilan), we compute metrics using VLMEvalKit. Since specialized OCR models typically do not report these metrics, we integrate their OCR outputs in our evaluation environment. Note that these benchmarks include not only pure OCR recognition but also understanding and key information extraction (KIE) tasks, where specialized OCR models generally underperform. For document understanding tasks, specialized OCR models cannot directly complete such tasks (metrics approach zero). To simulate real-world usage scenarios, we employ a two-stage pipeline: first using specialized OCR models for text extraction, then feeding the extracted text to Qwen3-4B LLM for answer generation.

\subsection{OCR-Specific Benchmarks}

\begin{table*}[htbp]
\centering
\small
\setlength{\tabcolsep}{4pt}
\begin{tabular}{l|c|cccccccc}
\toprule
\textbf{Model} & \textbf{OlmOCR} & \textbf{ArXiv} & \textbf{Old scans} & \textbf{Tables} & \textbf{Old} & \textbf{Headers \&} & \textbf{Multi} & \textbf{Long tiny} & \textbf{Base} \\
 & \textbf{Bench} & & \textbf{math} & & \textbf{scans} & \textbf{footers} & \textbf{column} & \textbf{text} & \\
\midrule
\multicolumn{10}{c}{\textbf{Pipeline}} \\
\midrule
PaddleOCR-VL & \textbf{80.0} & \textbf{85.7} & \textbf{71.0} & \textbf{84.1} & \textbf{37.8} & \textbf{97.0} & 79.9 & 85.7 & 98.5 \\
PaddleOCR-VL-1.5 & 79.1 & 84.6 & 63.8 & 81.8 & 36.9 & 95.5 & \textbf{80.9} & \textbf{90.5} & \textbf{98.9} \\
MonkeyOCR & 75.8 & 83.8 & 68.8 & 74.6 & 36.1 & 91.2 & 76.6 & 80.1 & 95.3 \\
\midrule
\multicolumn{10}{c}{\textbf{End-to-end Model}} \\
\midrule
MinerU2.5 & 75.2 & 76.6 & 54.6 & 84.9 & 33.7 & \textbf{96.6} & 78.2 & 83.5 & 93.7 \\
Dotsocr & 79.1 & \textbf{82.1} & 64.2 & \textbf{88.3} & 40.9 & 94.1 & 82.4 & 81.2 & 99.5 \\
PointsReader & 71.8 & 81.0 & 74.0 & 86.0 & 36.5 & 19.6 & \textbf{85.2} & \textbf{92.8} & 99.1 \\
DeepSeek-OCR & 77.2 & 77.2 & 73.6 & 80.2 & 33.3 & 96.1 & 66.4 & 79.4 & \textbf{99.8} \\
Qwen3-VL-4B & 79.2 & 81.9 & \textbf{76.6} & 79.3 & 40.7 & 88.8 & 78.8 & 88.0 & 99.1 \\
Qwen3-VL-2B & 70.7 & 78.1 & 72.5 & 69.3 & 38.0 & 51.2 & 76.1 & 82.1 & 98.6 \\
\textbf{Qianfan-OCR (Ours)} & \textbf{79.8} & 80.1 & 73.1 & 81.6 & \textbf{42.0} & 92.2 & 80.4 & 89.1 & 99.6 \\
\bottomrule
\end{tabular}
\caption{Detailed performance on OlmOCR Bench across document categories. Models are categorized into Pipeline (two-stage) and End-to-end architectures. Best results in each section are in \textbf{bold}.}
\label{tab:ocr_specific}
\end{table*}

\FloatBarrier

Table~\ref{tab:ocr_specific} presents detailed performance on OlmOCR Bench across document categories. Qianfan-OCR achieves the highest overall score (79.8) among end-to-end models, competitive with the top pipeline system PaddleOCR-VL (80.0), with strong scores on Base (99.6), Headers \& footers (92.2), and Old scans (42.0, best among end-to-end).

\begin{table*}[htbp]
\centering
\small
\setlength{\tabcolsep}{5pt}
\begin{tabular}{l|c|c|c|c|c|c}
\toprule
\textbf{Model} & \textbf{Overall} $\uparrow$ & \textbf{Text$^{\text{Edit}}$} $\downarrow$ & \textbf{Formula$^{\text{CDM}}$} $\uparrow$ & \textbf{Table$^{\text{TEDs}}$} $\uparrow$ & \textbf{Table$^{\text{TEDss}}$} $\uparrow$ & \textbf{R-order$^{\text{Edit}}$} $\downarrow$ \\
\midrule
\multicolumn{7}{c}{\textbf{Pipeline}} \\
\midrule
Marker-1.8.2 & 71.30 & 0.206 & 76.66 & 57.88 & 71.17 & 0.250 \\
MinerU2-pp & 71.51 & 0.209 & 76.55 & 70.90 & 79.11 & 0.225 \\
Dolphin & 74.67 & 0.125 & 67.85 & 68.70 & 77.77 & 0.124 \\
Dolphin-1.5 & 83.21 & 0.092 & 80.78 & 78.06 & 84.10 & 0.080 \\
PP-StructureV3 & 86.73 & 0.073 & 85.79 & 81.68 & 89.48 & 0.073 \\
MonkeyOCR-pro-1.2B & 86.96 & 0.084 & 85.02 & 84.24 & 89.02 & 0.130 \\
MonkeyOCR-3B & 87.13 & 0.075 & 87.45 & 81.39 & 85.92 & 0.129 \\
MonkeyOCR-pro-3B & 88.85 & 0.075 & 87.25 & 86.78 & 90.63 & 0.129 \\
MinerU2.5 & 90.67 & 0.047 & 88.46 & 88.22 & 92.38 & 0.044 \\
PaddleOCR-VL & 92.86 & 0.035 & 91.22 & 90.89 & 94.76 & 0.043 \\
PaddleOCR-VL 1.5 & \textbf{94.50} & \textbf{0.035} & \textbf{94.21} & \textbf{92.76} & \textbf{95.79} & \textbf{0.042} \\
\midrule
\multicolumn{7}{c}{\textbf{End-to-end Model}} \\
\midrule
OCRFlux & 74.82 & 0.193 & 68.03 & 75.75 & 80.23 & 0.202 \\
GPT-4o & 75.02 & 0.217 & 79.70 & 67.07 & 76.09 & 0.148 \\
InternVL3 & 80.33 & 0.131 & 83.42 & 70.64 & 77.74 & 0.113 \\
Qwen3-VL-2B & 80.58 & 0.103 & 82.32 & 69.71 & 74.14 & 0.113 \\
POINTS-Reader & 80.98 & 0.134 & 79.20 & 77.13 & 81.66 & 0.145 \\
olmOCR & 81.79 & 0.096 & 86.04 & 68.92 & 74.77 & 0.121 \\
InternVL3.5-241B & 82.67 & 0.142 & 87.23 & 75.00 & 81.28 & 0.125 \\
MinerU2-VLM & 85.56 & 0.078 & 80.95 & 83.54 & 87.66 & 0.086 \\
Nanonets-OCR-s & 85.59 & 0.093 & 85.90 & 80.14 & 85.57 & 0.108 \\
Qwen3-VL-4B & 86.78 & 0.055 & 86.55 & 79.29 & 84.26 & 0.084 \\
Qwen2.5-VL-72B & 87.02 & 0.094 & 88.27 & 82.15 & 86.22 & 0.102 \\
Gemini-2.5 Pro & 88.03 & 0.075 & 85.82 & 85.71 & 90.29 & 0.097 \\
dots.ocr & 88.41 & 0.048 & 83.22 & 86.78 & 90.62 & 0.053 \\
OCRVerse & 88.56 & 0.058 & 86.91 & 84.55 & 88.45 & 0.071 \\
Qwen3-VL-235B & 89.15 & 0.069 & 88.14 & 86.21 & 90.55 & 0.068 \\
Gemini-3 Pro & 90.33 & 0.065 & 89.18 & 88.28 & 90.29 & 0.071 \\
DeepSeek-OCR-v2 & 91.09 & 0.048 & 90.31 & 87.75 & 92.06 & 0.057 \\
\textbf{Qianfan-OCR (Ours)} & \textbf{93.12} & \textbf{0.041} & \textbf{92.43} & \textbf{91.02} & \textbf{93.85} & \textbf{0.049} \\
\bottomrule
\end{tabular}
\caption{Performance comparison on OmniDocBench. Models are categorized into Pipeline (two-stage) and End-to-end architectures. $\uparrow$ indicates higher is better, $\downarrow$ indicates lower is better. Best results in each section are in \textbf{bold}. Data source: \url{https://github.com/opendatalab/OmniDocBench}}
\label{tab:omnidocbench}
\end{table*}

Table~\ref{tab:omnidocbench} presents results on OmniDocBench v1.5. Among end-to-end models, Qianfan-OCR achieves the highest overall score of 93.12, surpassing DeepSeek-OCR-v2 (91.09), Gemini-3 Pro (90.33), and all other end-to-end models across every sub-metric. Notably, Qianfan-OCR also outperforms several pipeline systems including MinerU2.5 (90.67) and MonkeyOCR-pro-3B (88.85), narrowing the gap with the top pipeline system PaddleOCR-VL 1.5 (94.50).

\textbf{Analysis of Layout-as-Thought (Thinking Mode).} We additionally evaluate Qianfan-OCR with the Layout-as-Thought mechanism enabled at inference time (denoted Qianfan-OCR-think). Compared to the default mode, the thinking variant achieves an overall score of 92.64 vs.\ 93.12, with per-metric results: Text$^{\text{Edit}}$ 0.052 vs.\ 0.041, Formula$^{\text{CDM}}$ 91.92 vs.\ 92.43, Table$^{\text{TEDs}}$ 91.21 vs.\ 91.02, Table$^{\text{TEDss}}$ 94.03 vs.\ 93.85, and R-order$^{\text{Edit}}$ 0.051 vs.\ 0.049. While the aggregate score is slightly lower, the thinking variant shows improvements on table-related metrics (Table$^{\text{TEDs}}$: +0.19, Table$^{\text{TEDss}}$: +0.18). More importantly, per-sample analysis reveals that Layout-as-Thought provides targeted benefits on structurally complex documents but can be counterproductive on simpler ones.

\begin{figure}[htbp]
\centering
\includegraphics[width=0.95\linewidth]{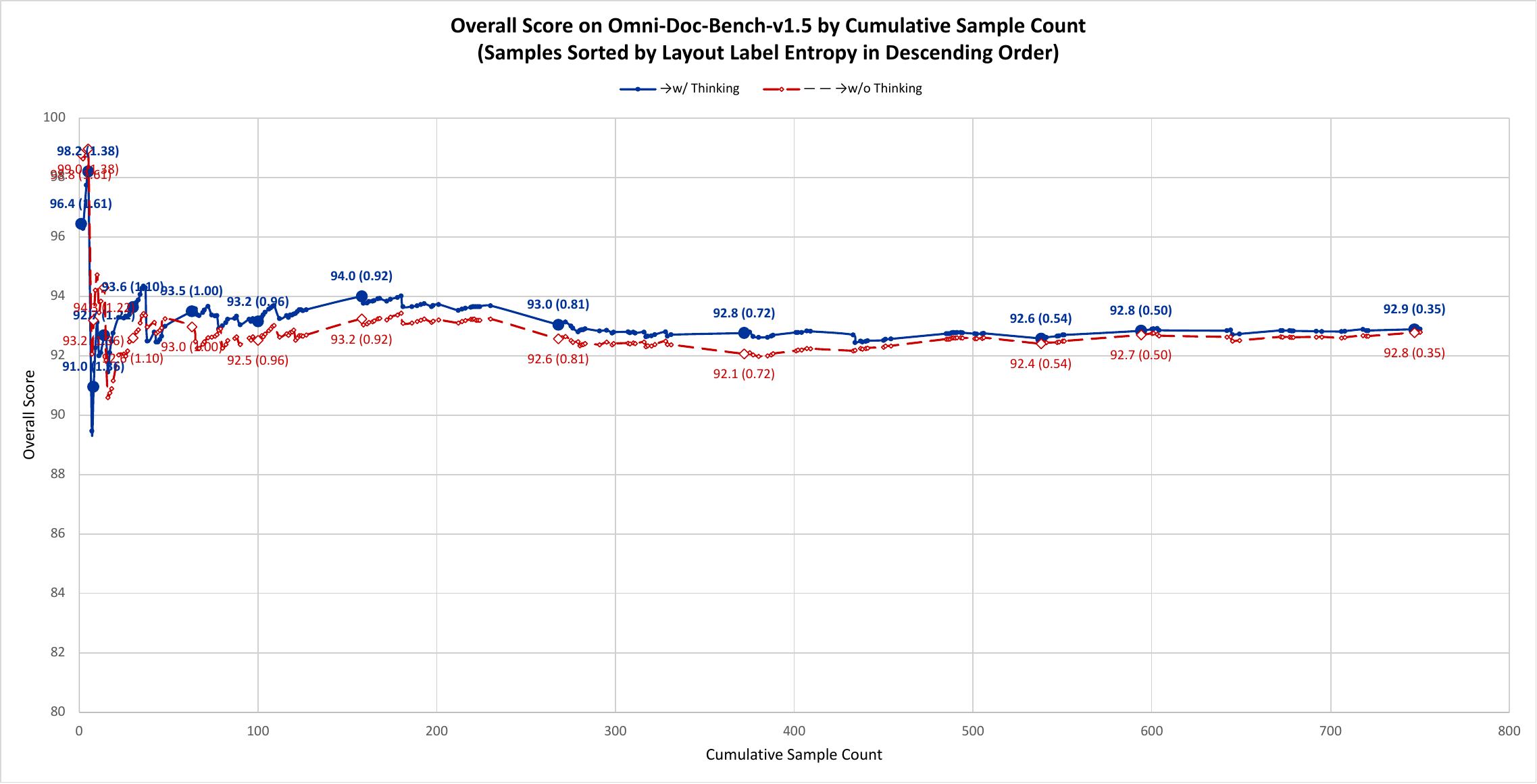}
\caption{Cumulative OmniDocBench v1.5 score with samples sorted by layout label entropy (descending). In the high-entropy region (left), enabling thinking provides a stable advantage. As lower-entropy samples are included, the gap narrows and eventually reverses, with the no-think mode achieving a higher total score overall.}
\label{fig:think_compare}
\end{figure}

To investigate this, we sort all OmniDocBench v1.5 samples by their layout label entropy in descending order and plot the cumulative score as samples are progressively included (Figure~\ref{fig:think_compare}). Layout label entropy measures the diversity of element types (text, table, formula, figure, etc.) on each page -- higher entropy indicates more heterogeneous layouts.

The results reveal a clear pattern: in the high-entropy region (left portion of the curve), where documents contain diverse element types such as mixed text, formulas, tables, and figures, enabling thinking provides a consistent score advantage. As lower-entropy samples are progressively included -- documents with more homogeneous layouts (e.g., pure text pages) -- the gap narrows and eventually reverses, with the no-think mode achieving a higher cumulative total. This indicates that Layout-as-Thought introduces unnecessary overhead on structurally simple documents, where explicit layout reasoning provides no additional benefit and may even interfere with direct recognition.

In practice, users should decide whether to enable the thinking mode based on the layout complexity of their target documents: for heterogeneous pages with mixed element types (exam papers, technical reports, newspapers), enabling thinking improves accuracy; for homogeneous documents (single-column text, simple forms), disabling thinking yields better results with lower latency.

\textbf{Illustrative Example.} Figure~\ref{fig:layout_example} presents a concrete example of Layout-as-Thought on a math exam paper with mixed content (text, formulas, geometric diagrams, and multi-column layout). When the \texttt{<think>} token is activated, the model first generates a structured layout analysis enumerating each document element in reading order:

\begin{figure*}[htbp]
\centering
\begin{minipage}[t]{0.48\textwidth}
\centering
\includegraphics[width=\linewidth]{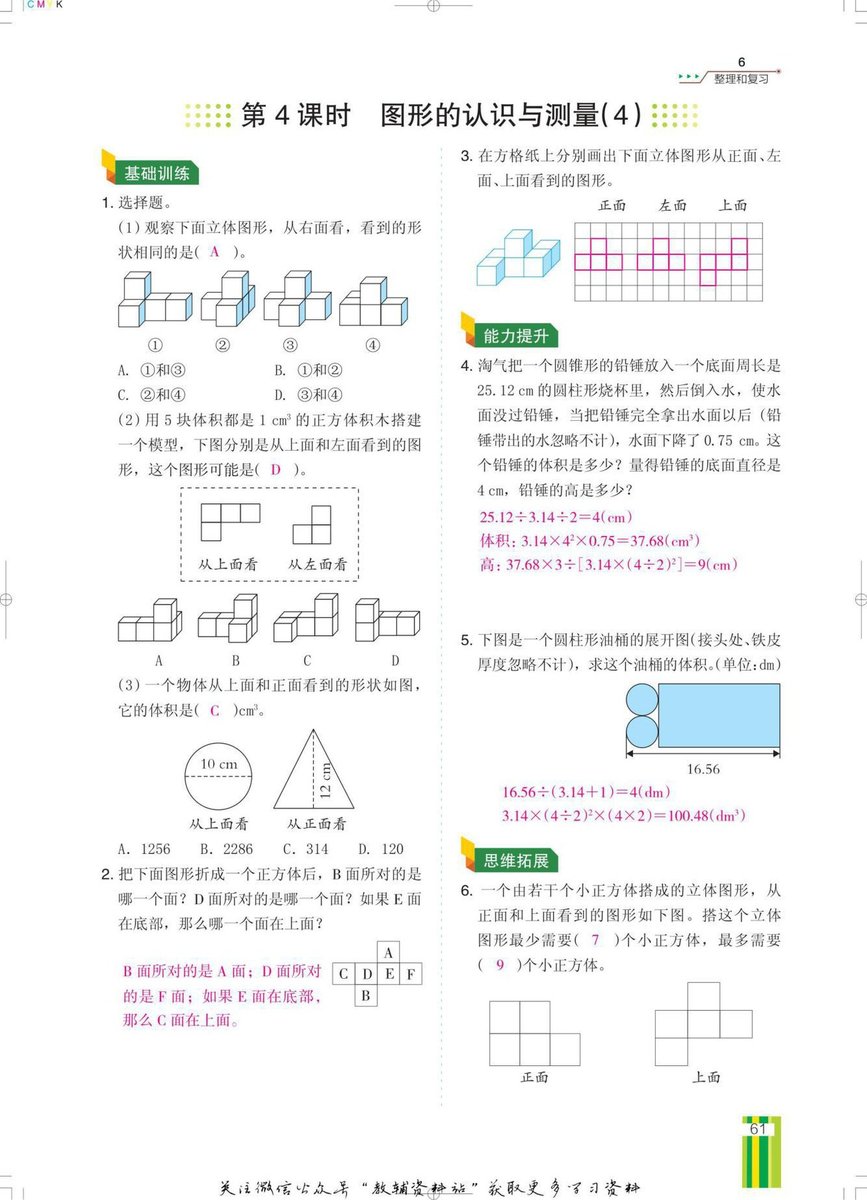}
\end{minipage}
\hfill
\begin{minipage}[t]{0.48\textwidth}
\centering
\includegraphics[width=\linewidth]{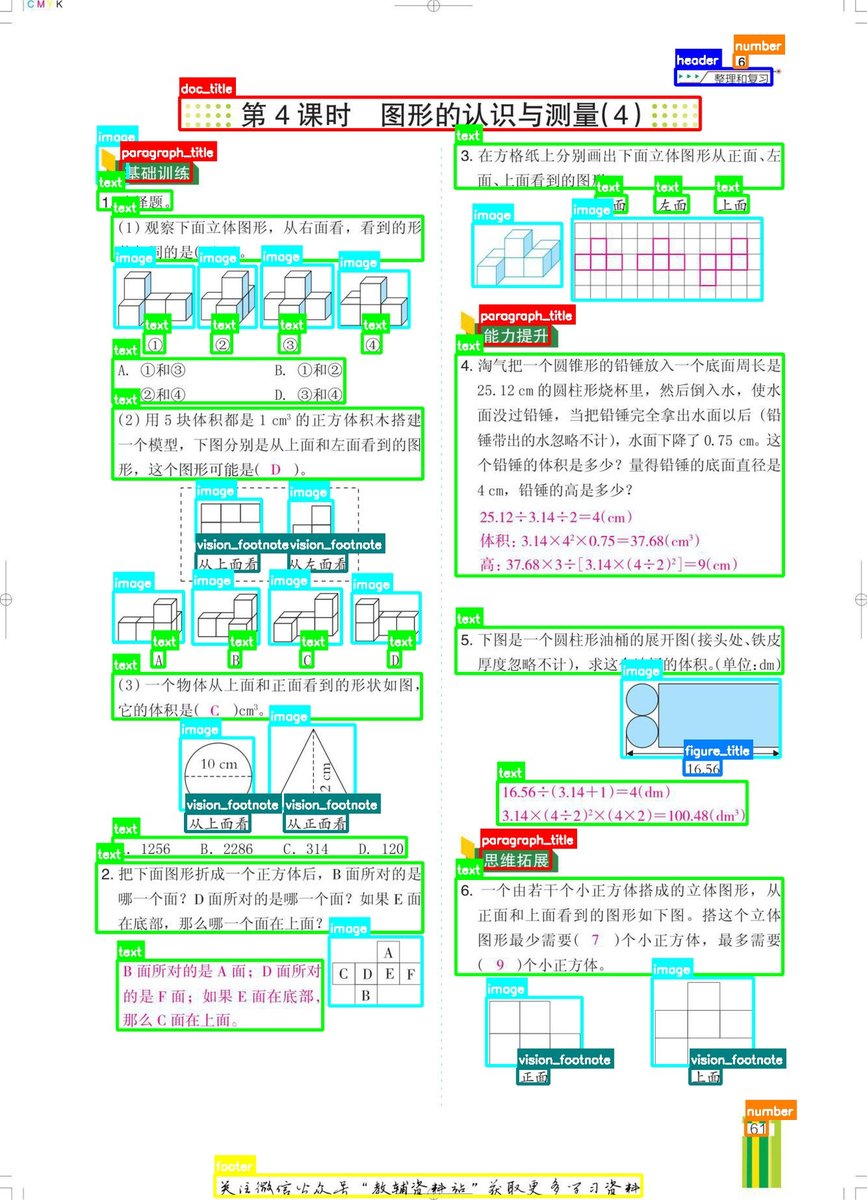}
\end{minipage}
\caption{Layout-as-Thought example on a math exam paper. \textbf{Left:} Original document image. \textbf{Right:} Visualization of bounding boxes generated during the thinking phase, with element types color-coded (e.g., \textcolor{red}{text}, \textcolor{cyan}{image}, \textcolor{orange}{paragraph\_title}, \textcolor{green}{vision\_footnote}).}
\label{fig:layout_example}
\end{figure*}

{\small
\begin{verbatim}
<think>
<box>[[<COORD_779>,<COORD_057>,<COORD_890>,<COORD_071>]]</box>
<label>header</label>
<brief>Section title indicating review and organization</brief>
<box>[[<COORD_207>,<COORD_081>,<COORD_808>,<COORD_108>]]</box>
<label>doc_title</label>
<brief>Lesson title: Graphical Recognition and Measurement (4)</brief>
<box>[[<COORD_139>,<COORD_134>,<COORD_221>,<COORD_151>]]</box>
<label>paragraph_title</label>
<brief>Subsection title: Basic Training</brief>
<box>[[<COORD_129>,<COORD_180>,<COORD_488>,<COORD_217>]]</box>
<label>text</label>
<brief>Asks to identify shapes seen from the right side...</brief>
<box>[[<COORD_132>,<COORD_222>,<COORD_224>,<COORD_273>]]</box>
<label>image</label>
<brief>A 3D arrangement of white cubes with light blue edges...</brief>
  ...  (60+ layout elements in reading order)
</think>
\end{verbatim}
}

Each entry contains three fields: a \texttt{<box>} with normalized bounding box coordinates, a \texttt{<label>} indicating the element type (from 25 layout categories including \textit{text}, \textit{image}, \textit{paragraph\_title}, \textit{display\_formula}, \textit{table}, \textit{vision\_footnote}, etc.), and a \texttt{<brief>} providing a concise content summary for text-type elements. The model then generates the final OCR response conditioned on this structured layout reasoning.

\textbf{Coordinate Special Tokens.} All bounding box coordinates are normalized to [0, 999] and represented as dedicated special tokens \texttt{<COORD\_0>} through \texttt{<COORD\_999>}. These tokens are introduced during Stage~3 continued pretraining alongside layout analysis data, enabling the model to learn spatial representations efficiently. Compared to encoding coordinates as plain digit sequences (e.g., ``779'' requires 3 tokens), each coordinate consumes only a single token, reducing the thinking output length by approximately 50\% and substantially decreasing inference latency for the layout reasoning phase. This compact representation is critical for practical deployment, as complex documents may contain 60+ layout elements in a single page.

\textbf{How Layout-as-Thought Guides Response Generation.} The structured thinking output benefits the final response in two key ways: (1) \textit{Element-type-aware generation}: by explicitly identifying element categories (formula, table, image, text), the model applies appropriate rendering formats in the response -- wrapping mathematical content in \$\$ ... \$\$ blocks, converting table structures to HTML, and inserting image placeholders at correct positions; (2) \textit{Reading-order-guided sequencing}: the thinking phase enumerates elements following the natural reading order of the document (handling multi-column layouts, interleaved figures, and footnotes), providing an explicit ordering signal that the response generation can follow to produce correctly sequenced output.

\subsection{General OCR Benchmarks}

Table~\ref{tab:general_ocr} shows performance on general OCR benchmarks covering diverse scenarios including scene text, documents, handwriting, and multilingual text. Qianfan-OCR achieves the highest OCRBench score (880) among all models and demonstrates strong performance on OCRBenchv2 (56.0 en / 60.77 zh), CCOCR-multilan (76.7), and CCOCR-overall (79.3). Compared with the similarly-sized general-purpose VLM Qwen3-VL-4B, Qianfan-OCR surpasses it on OCRBench (880 vs.\ 873) while showing a modest gap on OCRBenchv2 en (56.0 vs.\ 60.68), and leading on both CCOCR metrics (76.7 vs.\ 74.2 multilan, 79.3 vs.\ 76.5 overall). This trade-off is expected: Qianfan-OCR prioritizes specialized OCR capabilities (OmniDocBench 93.12, OlmOCR 79.8, both ranking first among all end-to-end models) while retaining competitive general performance without significant degradation.

\begin{table}[H]
\centering
\small
\setlength{\tabcolsep}{6pt}
\begin{tabular}{l|cccc}
\toprule
\textbf{Model} & \textbf{OCRBench} & \textbf{OCRBenchv2 (en/zh)} & \textbf{CCOCR-multilan} & \textbf{CCOCR-overall} \\
\midrule
Qianfan-OCR (Ours) & \textbf{880} & 56.0 / \textbf{60.77} & \textbf{76.7} & \textbf{79.3} \\
PaddleOCR-VL & 549 & 18.15 / 40.86 & 45.5 & 29.1 \\
MinerU2.5 & 331 & 17.09 / 40.03 & 43.2 & 24.4 \\
Dotsocr & 625 & 19.59 / 35.94 & 47.2 & 34.6 \\
MonkeyOCR & 655 & 21.78 / 38.91 & 43.8 & 35.2 \\
PointsReader & 620 & 23.40 / 40.39 & 43.5 & 35.0 \\
DeepSeek-OCR & 459 & 15.98 / 38.31 & 32.5 & 27.6 \\
Qwen3-VL-4B & 873 & \textbf{60.68} / 59.13 & 74.2 & 76.5 \\
Qwen3-VL-2B & 856 & 54.18 / 54.41 & 73.0 & 71.5 \\
\bottomrule
\end{tabular}
\caption{Performance on general OCR benchmarks. Qianfan-OCR achieves competitive scores while maintaining strong understanding capabilities (Table~\ref{tab:doc_understanding_comparison}).}
\label{tab:general_ocr}
\end{table}

Notable observations:
\begin{itemize}[noitemsep,topsep=0pt]
\item \textbf{OCRBench:} Qianfan-OCR scores 880, surpassing Qwen3-VL-4B (873) and ranking first among all models
\item \textbf{OCRBenchv2:} Qwen3-VL-4B leads on English (60.68 vs.\ 56.0), while Qianfan-OCR achieves the best Chinese text recognition (60.77), significantly outperforming all specialized OCR models on both languages
\item \textbf{CCOCR:} Qianfan-OCR leads on both CCOCR-multilan (76.7 vs.\ 74.2) and CCOCR-overall (79.3 vs.\ 76.5) over Qwen3-VL-4B, demonstrating strong multilingual and comprehensive OCR capabilities
\end{itemize}

\subsection{Document Understanding Benchmarks}

Document understanding benchmarks evaluate the model's ability to perform visual question answering, chart interpretation, and reasoning tasks requiring both accurate text perception and semantic comprehension.

Table~\ref{tab:doc_understanding_comparison} compares end-to-end models with two-stage OCR+LLM systems, where specialized OCR models first extract text and Qwen3-4B then generates answers from the extracted text.

\begin{table}[H]
\centering
\small
\setlength{\tabcolsep}{4pt}
\begin{tabular}{l|ccc|cccc}
\toprule
 & \multicolumn{3}{c|}{\textbf{End-to-End Models}} & \multicolumn{4}{c}{\textbf{Two-Stage OCR+LLM Systems}} \\
\cmidrule(lr){2-4} \cmidrule(lr){5-8}
\textbf{Benchmark} & \textbf{Qianfan} & \textbf{Qwen3} & \textbf{Qwen3} & \textbf{PaddleOCR} & \textbf{MinerU2.5} & \textbf{Dotsocr} & \textbf{DeepSeek OCR} \\
 & \textbf{-OCR} & \textbf{-VL-4B} & \textbf{-VL-2B} & \textbf{+Qwen3-4B} & \textbf{+Qwen3-4B} & \textbf{+Qwen3-4B} & \textbf{+Qwen3-4B} \\
\midrule
OCRVQA\_TESTCORE & \textbf{66.8} & 64.7 & 59.3 & 20.4 & 21.2 & 26.5 & 19.3 \\
TextVQA\_VAL & 80.0 & \textbf{81.8} & 79.9 & 21.4 & 16.1 & 28.0 & 40.0 \\
DocVQA & 92.8 & \textbf{94.9} & 92.7 & 59.8 & 62.0 & 67.1 & 55.9 \\
CharXiv\_DQ & \textbf{94.0} & 81.8 & 69.7 & 0.0 & 0.0 & 0.0 & 0.0 \\
CharXiv\_RQ & \textbf{85.2} & 48.5 & 41.3 & 0.0 & 0.0 & 0.0 & 0.0 \\
ChartQA\_TEST & \textbf{88.1} & 83.3 & 78.3 & 56.8 & 7.0 & 14.6 & 23.4 \\
ChartQAPro & \textbf{42.9} & 36.2 & 24.5 & 20.9 & 18.7 & 18.4 & 18.1 \\
ChartBench & \textbf{85.9} & 74.9 & 73.2 & 17.2 & 5.7 & 2.0 & 6.1 \\
\bottomrule
\end{tabular}
\caption{End-to-end models vs.\ two-stage OCR+LLM systems on document understanding benchmarks. Two-stage systems use Qwen3-4B as the downstream LLM. Best end-to-end results in \textbf{bold}.}
\label{tab:doc_understanding_comparison}
\end{table}

Qianfan-OCR achieves the best scores on six out of eight benchmarks, with particular strength on chart and academic reasoning tasks: CharXiv\_DQ (94.0), CharXiv\_RQ (85.2), ChartQA (88.1), ChartQAPro (42.9), ChartBench (85.9), and OCRVQA (66.8). On general document understanding benchmarks, Qwen3-VL-4B outperforms Qianfan-OCR on DocVQA (94.9 vs.\ 92.8) and TextVQA (81.8 vs.\ 80.0). This gap is consistent with the trend observed in Section~6.2: as a model designed for specialized OCR tasks, Qianfan-OCR accepts modest performance differences on general understanding benchmarks relative to same-size general VLMs, while achieving substantially stronger results on domain-specific OCR benchmarks (OmniDocBench, OlmOCR) and chart-related tasks where structural visual reasoning is critical.

The most striking result is the complete failure of all two-stage systems on CharXiv (0.0 on both DQ and RQ), where chart structures, axis relationships, and data point positions -- discarded during text extraction -- are essential for answering. The degradation extends to other chart benchmarks (ChartQA: 7.0--56.8 vs.\ 88.1; ChartBench: 2.0--17.2 vs.\ 85.9) and even text-heavy tasks like DocVQA (55.9--67.1 vs.\ 92.8--94.9), confirming that spatial and layout context provides value beyond what plain text can capture.

\subsection{Key Information Extraction Benchmarks}

We evaluate KIE performance on five public benchmarks as described in Section~\ref{sec:evaluation_framework}. Since specialized OCR models lack native KIE capabilities, we compare Qianfan-OCR with commercial models (Gemini-3.1-Pro, Gemini-3-Pro, Seed-2.0) and open-source models (Qwen3-4B-VL, Qwen3-VL-235B-A22B).

\begin{table}[H]
\centering
\small
\setlength{\tabcolsep}{6pt}
\begin{tabular}{l|cccccc}
\toprule
\textbf{Model} & \textbf{Overall} & \textbf{OCRBench} & \textbf{OCRBenchv2} & \textbf{OCRBenchv2} & \textbf{CCOCR} & \textbf{Nanonets} \\
 & \textbf{(Mean)} & \textbf{KIE} & \textbf{KIE (en)} & \textbf{KIE (zh)} & \textbf{KIE} & \textbf{KIE (F1)} \\
\midrule
\textbf{Qianfan-OCR (Ours)} & \textbf{87.9} & 95.0 & 82.8 & \textbf{82.3} & 92.8 & \textbf{86.5} \\
Qwen3-4B-VL & 83.5 & 89.0 & 82.1 & 71.3 & 91.6 & 83.3 \\
Qwen3-VL-235B-A22B & 84.2 & 94.0 & 85.6 & 62.9 & \textbf{95.1} & 83.8 \\
Gemini-3.1-Pro & 79.2 & \textbf{96.0} & \textbf{87.8} & 63.4 & 72.5 & 76.1 \\
Gemini-3-Pro & 77.0 & 93.5 & 87.1 & 49.6 & 72.7 & 82.1 \\
Seed-2.0 & 78.0 & 92.5 & 75.6 & 48.9 & 89.6 & 83.4 \\
\bottomrule
\end{tabular}
\caption{Performance comparison on Key Information Extraction (KIE) benchmarks (normalized scores). Qianfan-OCR achieves the highest overall mean score of 87.9 across five public KIE benchmarks, outperforming both state-of-the-art commercial models and similarly-sized open-source models.}
\label{tab:kie_benchmarks}
\end{table}

As shown in Table~\ref{tab:kie_benchmarks}, Qianfan-OCR achieves the highest overall mean score of 87.9 across five public KIE benchmarks, outperforming the similarly-sized Qwen3-4B-VL (83.5) by 4.4 points, the much larger Qwen3-VL-235B-A22B (84.2) by 3.7 points, and commercial large models by approximately 9--11 points. Qianfan-OCR leads on OCRBenchv2 KIE Chinese (82.3) and Nanonets KIE F1 (86.5), demonstrating strong advantages on Chinese document extraction and real-world evaluation scenarios. Notably, even Qwen3-VL-235B-A22B with over 50$\times$ more activated parameters achieves a lower overall score, primarily due to weak Chinese KIE performance (62.9 on OCRBenchv2 KIE zh). The Gemini-3 series achieves strong results on English KIE tasks (Gemini-3.1-Pro: OCRBench KIE 96.0, OCRBenchv2 KIE en 87.8), but drops sharply on Chinese benchmarks (OCRBenchv2 KIE zh: 63.4 and 49.6), indicating limited multilingual generalization. Seed-2.0 shows relatively weaker performance on OCRBenchv2 KIE tasks (en 75.6, zh 48.9) compared to other models. These results validate the effectiveness of Qianfan-OCR's end-to-end architecture for spatial reasoning and field association tasks across both Chinese and English scenarios.

\subsection{Inference Throughput}

Beyond accuracy, inference throughput is a critical factor for production deployment. Since two-stage pipeline systems (e.g., PaddleOCR-VL) involve heterogeneous components beyond the language model (layout detection, element-wise recognition, post-processing), raw model throughput alone is not a meaningful comparison metric. We therefore adopt \textbf{pages per second (PPS)} -- the number of complete document pages parsed per second -- as a holistic throughput measure that captures end-to-end system efficiency. All benchmarks are conducted on the OmniDocBench v1.5 dataset using a single NVIDIA A100 GPU with vLLM 0.10.2, consistent with the inference framework version reported in the PaddleOCR-VL technical report.

\begin{table}[H]
\centering
\small
\setlength{\tabcolsep}{6pt}
\begin{tabular}{l|ccc}
\toprule
\textbf{Model} & \textbf{Batch Size} & \textbf{\# Query} & \textbf{PPS} \\
\midrule
PaddleOCR-VL$^\dagger$ & -- & -- & 1.224 \\
MinerU 2.5$^\dagger$ & -- & -- & 1.057 \\
MonkeyOCR-pro-1.2B$^\dagger$ & -- & -- & 0.673 \\
Dots OCR$^\dagger$ & -- & -- & 0.352 \\
\midrule
Qianfan-OCR (W16A16) & 512 & 512 & 0.503 \\
Qianfan-OCR (W8A8) & 512 & 512 & \textbf{1.024} \\
\bottomrule
\end{tabular}
\caption{Inference throughput comparison measured in pages per second (PPS) on OmniDocBench v1.5 with a single A100 GPU. $^\dagger$Results from PaddleOCR-VL technical report.}
\label{tab:throughput}
\end{table}

As shown in Table~\ref{tab:throughput}, despite having a 4B-parameter language model backbone -- substantially larger than the detection and recognition modules in pipeline systems -- Qianfan-OCR with W8A8 quantization achieves 1.024 PPS, comparable to PaddleOCR-VL (1.224 PPS) and exceeding MonkeyOCR-pro-1.2B (0.673 PPS) and Dots OCR (0.352 PPS).

This competitive throughput is attributable to two architectural advantages of the end-to-end approach:

\begin{itemize}[noitemsep,topsep=0pt]
\item \textbf{GPU-centric computation:} Two-stage pipeline systems rely on CPU-based layout analysis (detection, NMS, rule-based assembly) as a prerequisite for GPU-based recognition. Under high concurrency, the CPU stage becomes a bottleneck that throttles GPU utilization -- a problem that worsens with more powerful GPUs. Qianfan-OCR processes entire pages through the GPU with minimal CPU involvement, avoiding this bottleneck entirely.
\item \textbf{Efficient batching:} End-to-end models accept whole-page images that can be resized to uniform dimensions, enabling large-batch GPU inference with well-aligned memory access patterns. In contrast, pipeline systems process variable numbers of cropped regions per page, leading to irregular batch sizes and fragmented GPU utilization.
\item \textbf{Lower deployment complexity:} Pipeline systems such as PaddleOCR-VL require asynchronous orchestration of data loading, layout analysis, and LLM inference stages, demanding careful tuning of per-stage concurrency, queue depths, and resource allocation to achieve optimal throughput. Qianfan-OCR reduces this to a standard single-model serving problem (e.g., a single vLLM instance), significantly lowering deployment effort and performance tuning cost.
\end{itemize}

With AWQ quantization at W8A8 precision, Qianfan-OCR achieves a 2$\times$ speedup over the W16A16 baseline (1.024 vs.\ 0.503 PPS) with negligible accuracy degradation, making it a practical choice for high-throughput document processing pipelines.

\section{Limitations and Future Work}

Qianfan-OCR represents an early exploration of unified end-to-end document intelligence with several limitations that warrant further investigation.

\textbf{Layout-as-Thought.} The current Layout-as-Thought mechanism has only been validated on OmniDocBench v1.5 for document parsing, where it shows targeted benefits on structurally complex pages. Its effectiveness on other tasks -- such as key information extraction, document QA, and chart understanding -- remains unexplored. The current implementation generates layout bounding boxes, labels, and brief text descriptions in a relatively rigid format via supervised fine-tuning. Future work should integrate these layout elements more naturally into the reasoning process, allowing the model to flexibly invoke spatial reasoning when needed rather than producing a fixed-format layout dump. Reinforcement learning is a promising direction to achieve this: by optimizing layout generation based on downstream task rewards, the model can learn to produce task-adaptive layout reasoning that selectively emphasizes relevant structural information, ultimately strengthening reasoning capabilities across diverse document intelligence scenarios.

\textbf{Performance Ceiling.} As a pioneering attempt at end-to-end OCR, the ultimate performance ceiling of purely end-to-end architectures remains an open question -- future work should systematically explore architectural innovations, training strategies, and data scaling laws to determine whether end-to-end models can fully match or surpass heavily optimized pipeline systems.

\textbf{Deployment Efficiency.} Although W8A8 quantization enables competitive throughput on GPU (Section~6.5), Qianfan-OCR's 4B parameter footprint limits deployment in resource-constrained environments such as edge devices and CPU-only servers -- future work should explore knowledge distillation and pruning to develop compact variants (1B--2B parameters) suitable for broader deployment scenarios. Beyond these core challenges, Qianfan-OCR exhibits limitations in video OCR, 3D text on curved surfaces, and highly stylized artistic handwriting, presenting promising directions for extending the unified architecture.

\section{Conclusion}

We present Qianfan-OCR, a 4B-parameter end-to-end model that unifies text recognition, layout analysis, and semantic understanding within a single vision-language architecture. Our key contributions include: (1) achieving state-of-the-art results among end-to-end models on OmniDocBench v1.5 and OlmOCR Bench, demonstrating that end-to-end architectures can be competitive with pipeline systems on recognition accuracy; (2) introducing Layout-as-Thought, a mechanism that integrates layout reasoning as optional chain-of-thought, enabling the model to dynamically invoke structural analysis for complex documents; (3) providing empirical evidence that two-stage OCR+LLM pipelines degrade substantially on tasks requiring spatial and visual reasoning, with zero accuracy on chart interpretation benchmarks where layout information is essential.

These results suggest that for document intelligence tasks requiring joint visual and textual understanding, preserving visual context throughout the processing pipeline offers significant advantages over text-only intermediate representations.

The model is publicly accessible through Baidu AI Cloud Qianfan platform at \url{https://github.com/baidubce/Qianfan-VL}.

\section*{Acknowledgments}

We thank the Baidu AI Cloud team for infrastructure support, the Baige and Kunlun teams for AI infrastructure assistance, and all contributors to the QianFan platform. We are deeply grateful to the operations, storage, and network teams for maintaining the stability of the P800 clusters. Special thanks to our annotation teams and quality assurance engineers for their meticulous work in data validation.

\clearpage
\section*{Contributors}

\textbf{Core Contributors} \\
Daxiang Dong, Mingming Zheng, Dong Xu, Chunhua Luo, Bairong Zhuang, Yuxuan Li

\vspace{0.5em}
\textbf{Contributors} \\
Ruoyun He, Haoran Wang, Wenyu Zhang, Wenbo Wang, Yicheng Wang, Xue Xiong, Ayong Zheng, Xiaoying Zuo, Ziwei Ou, Jingnan Gu, Quanhao Guo

\vspace{0.5em}
\textbf{Project Sponsors} \\
Jianmin Wu, Dawei Yin, Dou Shen

\bibliography{qianfan_ocr}

\end{document}